\documentclass[a4paper,fleqn]{cas-dc}
\usepackage[numbers]{natbib}
\usepackage{siunitx}

\def\tsc#1{\csdef{#1}{\textsc{\lowercase{#1}}\xspace}}
\tsc{WGM}
\tsc{QE}
\tsc{EP}
\tsc{PMS}
\tsc{BEC}
\tsc{DE}

\DeclareUnicodeCharacter{2212}{-}
\DeclareUnicodeCharacter{03B5}{-}
\begin{document}

\let\WriteBookmarks\relax
\def\floatpagepagefraction{1}
\def\textpagefraction{.001}
\shorttitle{Skeleton-based Hand Gesture Recognition in the Wild}

\title [mode = title]{SHREC 2021: Track on Skeleton-based Hand Gesture Recognition in the Wild}                      


\author[1]{Ariel Caputo}
\author[1]{Andrea Giachetti}
\author[1]{Simone Soso}
\author[1]{Deborah Pintani}
\author[2]{Andrea D'Eusanio}
\author[2]{Stefano Pini}
\author[3]{Guido Borghi}
\author[2]{Alessandro Simoni}
\author[2]{Roberto Vezzani}
\author[2]{Rita Cucchiara}
\author[4]{Andrea Ranieri}
\author[4]{Franca Giannini}
\author[4]{Katia Lupinetti}
\author[4]{Marina Monti}
\author[5]{Mehran Maghoumi}
\author[5]{Joseph J. LaViola Jr}
\author[6]{Minh-Quan Le}
\author[6]{Hai-Dang Nguyen}
\author[6]{Minh-Triet Tran}

\address[1]{Department of Computer Science, University of Verona, Italy}
\address[2]{ Università di Modena e Reggio Emilia, Dipartimento di Ingegneria "Enzo Ferrari"}
\address[3]{Dipartimento di Informatica - Scienza e Ingegneria}
\address[4]{Istituto di Matematica Applicata e Tecnologie  Informatiche "Enrico Magenes"}
\address[5]{University of Central Florida, Department of Electrical Engineering and Computer Science}
\address[6]{University of Science, Ho Chi Minh City, Vietnam}












\begin{abstract}
Gesture recognition is a fundamental tool to enable novel interaction paradigms in a variety of application scenarios like Mixed Reality environments, touchless public kiosks, entertainment systems, and more. Recognition of hand gestures can be nowadays performed directly from the stream of hand skeletons estimated by software provided by low-cost trackers (Ultraleap) and MR headsets (Hololens, Oculus Quest) or by video processing software modules (e.g. Google Mediapipe). Despite the recent advancements in gesture and action recognition from skeletons, it is unclear how well the current state-of-the-art techniques can perform in a real-world scenario for the recognition of a wide set of heterogeneous gestures, as many benchmarks do not test online recognition and use limited dictionaries. This motivated the proposal of the SHREC 2021: Track on Skeleton-based Hand Gesture Recognition in the Wild. For this contest, we created a novel dataset with heterogeneous gestures featuring different types and duration. These gestures have to be found inside sequences in an online recognition scenario. This paper presents the result of the contest, showing the performances of the techniques proposed by four research groups on the challenging task compared with a simple baseline method.
\end{abstract}




\maketitle

\section{Introduction}
The recognition of gestures based on hand skeleton tracking is becoming the default interaction method for the new generation of Virtual Reality (VR) and Mixed Reality (MR) devices like Oculus Quest and Microsoft Hololens, implementing specific, advanced solutions \cite{han2020megatrack,ungureanu2020hololens}.
Low-cost hand tracking devices with good performances are available since 2010 \cite{weichert2013analysis} and are used in several application domains and research works.
Real-time hand pose tracking is now possible from single-camera input using Google tools \cite{zhang2020mediapipe}.
It is, therefore, extremely likely that most of the future hand gesture recognition tools will work directly on the 
hand skeleton poses and not on RGB or depth images.
These facts strongly motivate research efforts aimed at the development of such tools.
In practical application scenarios, these gesture recognizers need to work in real-time and to be able to detect and correctly label gestures "in the wild" within a continuous sequence of hand movements.

Several methods have been recently proposed in the literature for the skeleton-based gesture recognition task. However, as pointed out in \cite{sfinge3d},
current available benchmarks that focus on online-recogntion scenarios are limited.
Many of them do not test recognizers in an online setting or evaluate the methods on limited vocabularies and not including many gesture types.
Hand gestures, in fact, can be classified into different types according to their distinctive features.
Some gestures are static, characterized by keeping a fixed hand pose for a minimum amount of time. Others are dynamic and characterized by a single trajectory with the hand pose that does not change or it is not semantically relevant. Others are dynamic and characterized not only by a global motion, but also by the evolution of fingers' articulation over time.

Previous contests organized on skeleton-based gesture recognition were limited to offline recognition (SHREC’17 Track: 3D Hand Gesture Recognition Using a Depth and Skeletal Dataset \cite{de2017shrec}) or featured a very limited dictionary of gestures (SHREC 2019 track on online gesture detection \cite{udeepgru}).

\begin{figure*}[ht]
	\centering
	\includegraphics[width=1\linewidth]{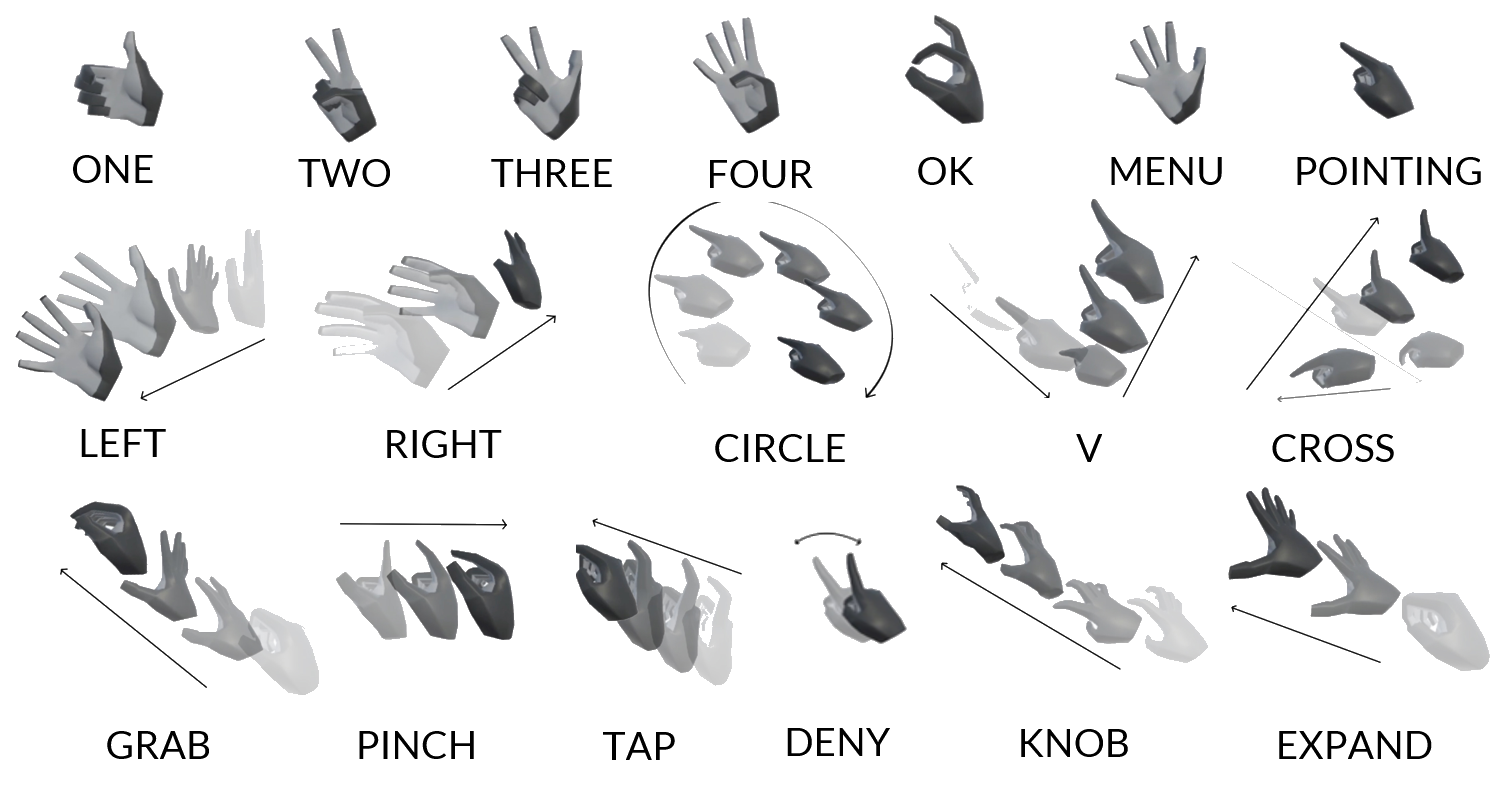}
	\caption{Gesture templates for the different classes. Top row: static gestures. Middle row: coarse dynamic gestures. Bottom row: fine dynamic gestures.}
	\label{fig:gestures}
\end{figure*}

For this reason, we created a novel dataset including 18 gesture classes belonging to different types. A subset of 7 classes are static, characterized by a hand pose kept fixed for at least one second (One, Two, Three, Four, OK, Menu, Pointing).
The remaining ones are dynamic, 5 coarse, characterized by a single global trajectory of the hand (Left, Right, Circle, V, Cross) and 6 fine, characterized by variations in the fingers' articulation (Grab, Pinch, Tap, Deny, Knob, Expand). 
Figure \ref{fig:gestures} shows the gestures' templates.
A peculiarity of the data collected is that the gestures are executed within long sequences of hand gesticulation, as they were captured during generic user interaction. 

Given the dataset, we proposed an online recognition task within the Eurographics SHREC 2021 framework. This paper reports on the outcomes of the contest's result. The paper is organized as follows: 
Section \ref{sec:dataset} presents the novel dataset, Section \ref{sec:task} the proposed task and the evaluation method, Section \ref{sec:parti} presents the groups participating in the contest and the methods proposed together with a baseline method.

\section{Dataset creation}
\label{sec:dataset}
The dataset created for the contest is a collection of 180 gesture sequences. Each sequence, captured using a Leap Motion Device, features either 3, 4, or 5 gestures interleaved with non-significant gesticulation. The dictionary used consists of 18 classes of gestures, each appearing in the dataset an equal amount of time (40 occurrences per class).

Gestures were performed by five different subjects in pre-determined sequences. The execution of the dictionary gestures followed specific templates shown in advance. Non-significant gesticulation was limited to a  restricted set of allowed movements to avoid biases on false detections.

We designed a randomized set of 36 sequences (12 with 3 gestures, 12 with 4 gestures, and 12 with 5 gestures). Each set thus includes 8 samples of the 18 gesture classes (a total of 144 gesture samples across the 36 sequences). Each subject recorded such a set of sequences for a total of 180 sequences with 720 gestures.

Three of the acquired sets (108 sequences with 24 samples of each gesture class) were given to the participants as the training set, with the associated annotations of gestures begin/end and labels.
The remaining two (72 sequences with 16 occurrences of each gesture class) were given to the participants as the test set without associated labels. Gestures in the test sets were performed by subjects not involved in the training set creation.

We performed the acquisition of the sequences using a simple Unity app running on a desktop PC. Subjects had to wear a headband with the leap motion device mounted over simulating sensor mounted on VR/AR glasses. The graphical interface of the app suggested the randomized temporal sequences of the gestures.

The accurate annotation of the timestamps of the start and end of the gestures was subsequently manually performed with another specific Unity application. In particular, the annotation tool allows to navigate the sequences, showing frame-by-frame the hand's movement. The application then allows to mark frames as the beginning or end of a gesture.

\section{Task proposed and evaluation}
\label{sec:task}
The goal of the participants was to detect correctly the gestures included in the sequence with an online detection approach.
The gesture database captured was split as described into a training set with associated annotations of gesture time stamp and labels, that could be used to train the detection algorithms and a test set with no annotations available.
Participants had to provide a list of the gestures detected in the test set with associated labels, start time stamps and end timestamps.

The results have been evaluated using different metrics.
First, we use the Jaccard Index as proposed in other online gesture recognition contests \cite{wan2016chalearn,zhang2018egogesture} to measure the average relative overlap between the ground truth and the predicted label sequences for the gesture sequences recording.
If $GT_{s,i}$ is a vector corresponding to the time sampling of the gesture with label $i$ filled with ones where the gesture is actually performed and zeros elsewhere, and $P_{s,i}$  is the corresponding prediction, the index is given by:

\begin{equation}
    JI_{s,i}=\frac{GT_{s,i}\cap P_{s,i}}{GT_{s,i}\cup P_{s,i}}
\end{equation}

Following our previous SHREC contest on Online Gesture recognition \cite{udeepgru}, we defined specific metrics to obtain an estimation of how the specific methods can measure the performances of a recognizer in a typical use context, where we are interested in detecting correctly gestures with a short delay after their execution and avoid false detections. We, therefore, counted the "detection rate" in the test data, e.g. the percentage of predicted gestures (of each class) corresponding to ground truth ones correctly detected.
A prediction is considered corresponding to the ground truth if it has a temporal intersection ratio with the ground truth one higher than $0.5$ and the same class label.
We then measure also the false-positive ratio, i.e. the ratio between the number of gestures (of a particular class) predicted and not corresponding to ground truth ones divided by the total number of gestures of that class in the sequences. 

\section{Participants and Methods}
Five research groups were registered for the contest and sent results, but one retired after the evaluation. Each group sent up to three annotation files that were obtained with different methods or parameters' settings. The methods are described in the following subsection, together with the simple technique that we used as baseline. 

\label{sec:parti}
\subsection{Baseline: Dissimilarity-based Classification}
As a baseline method, we customized an algorithm used in \cite{sfinge3d} based on class-specific binary classifiers trained with dissimilarity features and a sliding window approach for online detection. 
We created a gesture dictionary with the segmented labeled intervals cropped from the training sequence and a set of "non-gesture" examples randomly cropped from non-labeled parts of the sequences. Each gesture has been re-sampled to 20 time steps.
Half of the data have been used as the representation set for the dissimilarity vectors estimation.

\begin{figure}[t]
    \centering
    \includegraphics[width=0.8\linewidth]{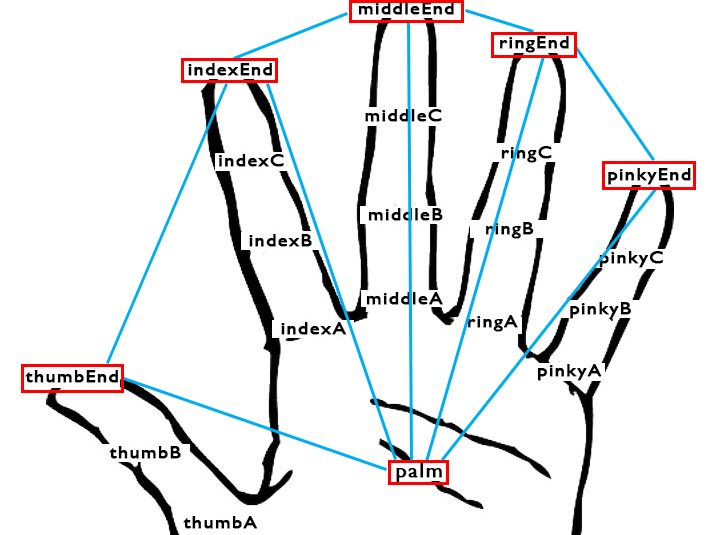}
    \caption{Keypoints of the hand's skeleton (red) used to calculate the features for the dissimilarity-based method (4.1)}
    \label{fig:keypoints}
\end{figure}

Given the re-sampled gestures, we define four sets of dissimilarity vectors:
\begin{itemize}
    \item palm trajectory dissimilarity: for each gesture of the representation set, we estimate the sum of the Euclidean distances of the corresponding points in query one. Given N  gestures in the representation set, we get N features for the query gesture descriptor.
    \item hand articulation dissimilarity: first we estimate the evolution of distances between adjacent fingertips and between fingertips and the palm keypoint (Fig~\ref{fig:keypoints}). We then calculate 9 dissimilarity components as the sums over corresponding time samples of the differences between the values of the 9 distances in the query and in the representation set gestures.
    \item palm trajectory length dissimilarity:  the difference in length between the query gesture and the representation set gestures. 
    \item palm velocity dissimilarity: the sum of the differences of the velocity magnitude samples at corresponding time steps,
\end{itemize}

For each gesture class, we train class-specific linear SVM classifiers using binary labels (gesture vs non-gesture).

During the online gesture recognition, we use a sliding window approach. To detect gestures at time $t$ we give as input for each trained classifier, the hand pose samples cropped in the window covering the interval $[t-l(c), t]$, where $l(c)$ is the average duration, in frames, of the gestures of class $c$ in the training set, re-sampled in 20 steps.
Windows are sampled every 6 frames of the original sequence. If a gesture is detected, we assign it a duration equal to $l(c)$. If a gesture is detected in multiple consecutive frames the predictions are merged and the estimated duration is incremented by the number of consecutive detections multiplied by the windows sampling step.

\subsection{Group 1: Transformer Network based method}
\subsubsection{Method Description}
Group 1 proposed a dynamic gesture recognition system based on the Transformer model~\cite{vaswani2017attention}.
The framework is based on different type and combinations of features.
The first group of features are provided by the \textit{Leap Motion SDK} and consist in the 3D position of the hand joints. 
They enrich this features computing the joint velocity and acceleration.
At time $t$, given the sequence of the $3$D position of the \textit{i}-th joint $J_i^t = (x_i^t,\; y_i^t,\; z_i^t)$, speed $\mathbf{s}$ and acceleration $\mathbf{a}$ are computed following these formulas:

\begin{equation}
\begin{aligned}
\textbf{s}_i^t = \big[ x_i^{t} - x_i^{(t-1)},\;\; y_i^{t} - y_i^{(t-1)},\;\; z_i^{t} - z_i^{(t-1)}  \big] \\
\textbf{a}_i^t = \big[ x_i^{t} - 2x_i^{(t-1)} + x_i^{(t-2)},\;\; y_i^{t} - 2y_i^{(t-1)} + y_i^{(t-2)},\nonumber\\ \;\; z_i^{t} - 2z_i^{(t-1)} + z_i^{(t-2)}  \big]
\end{aligned}
\end{equation}

Moreover, at time $t$ the joint-to-joint 3D distances expressed as a matrix $D$ of size $ 3 \times N \times N$, where $N$ is the total number of the hand joints, is also computed. 
Each element $d_{k,j,z} \in D$ is computed as:
\begin{equation}
    d_{k,j,z} = \sqrt{(J^t_{k,z} - J^t_{j,z})^2}, \quad k,j \in N, \quad z \in [0, 3]
\end{equation}
Therefore, the final dimension of the input feature vector depends on the types of the feature used.
A feature vector composed by the hand features provided by the \textit{Leap Motion} device and the speed and acceleration has size of $240$. 
Including the 3D position and rotation of each hand joint the feature vector has a size of $640$. 


As pre-processing, Group 1 tested two different normalization techniques. In the first one, they normalized the joint positions in a per-instance manner to obtain a per-axis zero mean and unit variance. In the second one, they additionally divided each joint position by the hand ``size'', set as the distance between the joints \textit{indexA} and \textit{pinkyA}, before the zero mean and unit variance normalization.

At inference time, the network predicts the current gesture on a single-step time scale and a \textit{Finite State Machine} (FSM) is used to detect the beginning and the end of each gesture.
Group 1 provided three different version of the presented algorithm: in the first and second cases, the feature vector is composed of the hand feature and speed and acceleration and then normalized using the two operations above mentioned. In the third case, all types of features normalized with zero mean and unit variance are included.

\subsubsection{Model Architecture}
The proposed method is composed of a transformer module~\cite{vaswani2017attention} (with internal dropout), followed by a fully connected layer applied to each step output which predicts the gesture class (including the class ``no-gesture'').

Formally, the model can be defined as:
\begin{equation}
    Y(\textbf{x}) = \text{F}(\text{Encoders}(\textbf{x} + PE))
\end{equation}
where $\text{F}(\cdot)$ corresponds to the fully connected layer that performs the gesture classification and the following softmax layer, applied to each time step $x \in \textbf{x}$, $\text{Encoders}(\cdot)$ represents a sequence of $6$ transformer encoders $E$, defined in the following, and PE is the Positional Encoding~\cite{vaswani2017attention}, used to encode the temporal information into the sequence.
Thus, $Y(\textbf{x})$ is a vector containing a probability distribution over $n$ gesture classes for each time step included in $\textbf{x}$.

Each transformer encoder is defined as
\begin{equation}
    E(x) = \text{Norm}(x + \text{FC}(\text{mhAtt}(x)))
\end{equation}
where $\text{Norm}(\cdot)$ is a normalization layer, $\text{FC}(\cdot)$ are two fully connected layers with $1024$ units, followed by a ReLU activation function.
The multi-head attention block $\text{mhAtt}$ is the self-attention layer defined as
\begin{equation}
    \text{mhAtt}(x) = (\, \text{Att}_1(x) \oplus \ldots \oplus \text{Att}_8(x) \, ) \, W^O
\end{equation}
where
\begin{equation}
    \text{Att}_i(x) = \text{softmax} \left(  \frac{Q_i \, K_i}{\sqrt{d_k}}  \right) \, V_i
\end{equation}
Here, $Q_i = x W^Q_i$, $K_i = x W^K_i$, $V_i = x W^V_i$ are independent linear projections of $x$ into a $64$-d feature space, $d_k = 64$ is a scaling factor corresponding to the feature size of $K_i$, $\oplus$ is the concatenation operator and $W^O$ is a linear projection from and to a $512$-d feature space.

\subsubsection{Training}
Since in the training data the large majority of frames are labelled as ``no-gesture'', Group 1 uses the \textit{Focal} loss~\cite{lin2017focal} which has been shown to handle well unbalanced training datasets. 
In this case, they empirically verified that the recognition accuracy was higher using this loss against the standard \textit{Categorical Cross Entropy} loss.
We train the model using \textit{Adam}~\cite{kingma2014adam} as optimizer with a learning rate $0.0001$, weight decay $0.0001$, and internal dropout set to $0.5$.
The features are fed to model which is trained with sliding windows of $10$ time steps.
We train the network on the whole training dataset for $5$ epochs.
The network hyper-parameters has been chosen with a k-fold cross validation (using $k=9$).

\begin{figure}[t]
    \centering
    \includegraphics[width=1\linewidth]{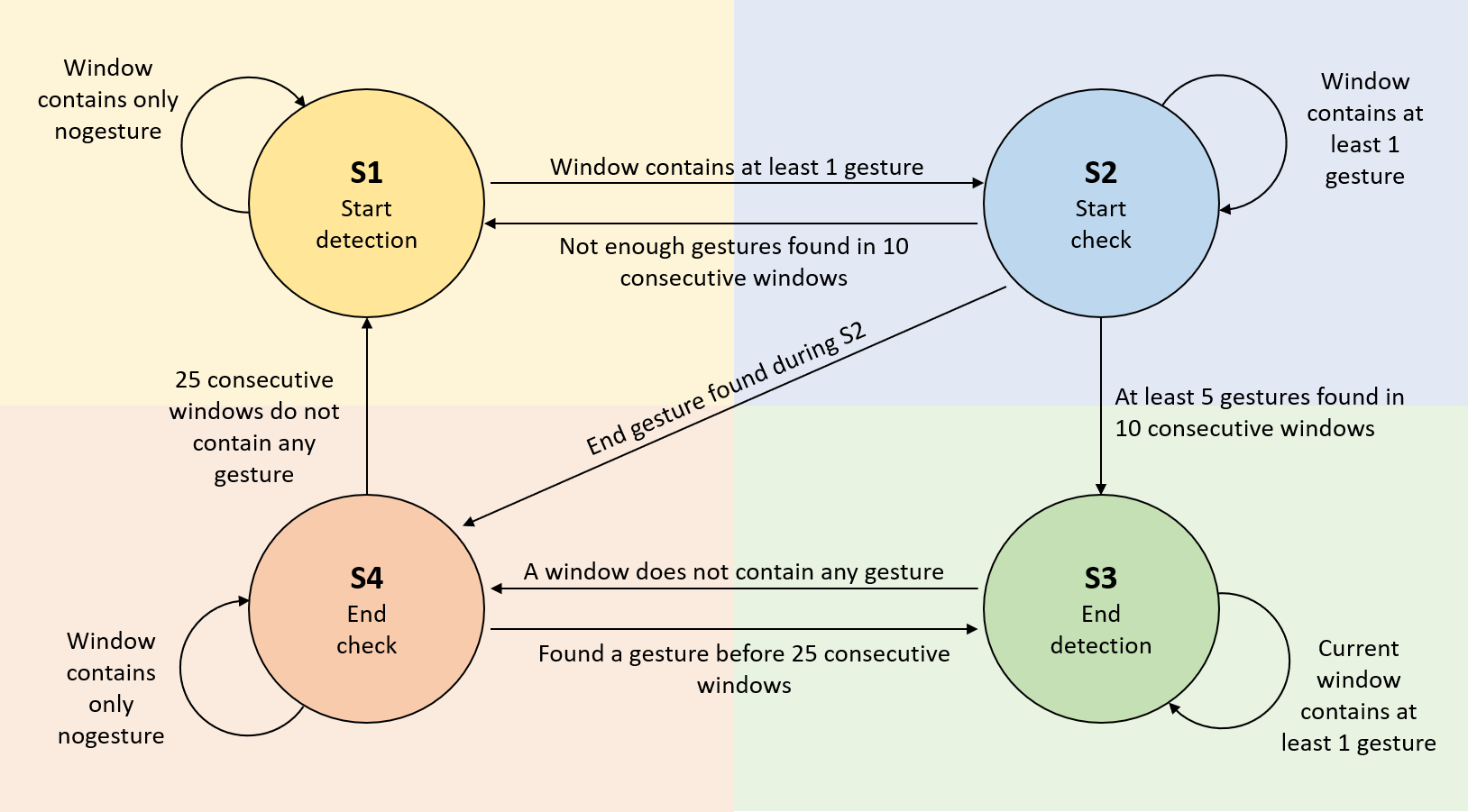}
    \caption{Visual description of the Finite State Machine implemented for the online gesture recognition.}
    \label{fig:fsm}
\end{figure}

\subsubsection{Online Detection}
During the testing phase, the proposed system receives one frame per time. 
Therefore, in order to detect the beginning, the end and the class of a gesture, Group 1 implemented a \textit{Finite State Machine} (FSM).
The input is represented by a buffer, i.e. a FIFO stack, with a size of $10$ frames.
The FSM with $4$ states, as depicted in Figure~\ref{fig:fsm} and it is running when the buffer is full of frames.

In the first state, the beginning of a gesture is detected. 
Each frame is classified by the proposed Transformer architecture and then, if even just a single frame is classified as gesture, the current state of the FSM is increased.

In the second state, a check on the beginning of the gesture is conducted. 
If not enough gestures are found in 10 consecutive windows, the FSM returns in the initial state. 
If at least 5 gestures are detected in 10 consecutive windows, the FSM goes in the third state.
If the end of a gesture is found, the FSM passes directly on the fourth state. 
This last case is related to the possible presence of gestures with a very limited duration.

In the third state, the end part of the gesture is detected.
The end of a gesture is an entire windows that does not contain any frame classified as gesture. 
In that case, the FSM passes to the fourth state.

In the fourth state, the end of the gesture is verified.
Indeed, only if $25$ consecutive windows do not contain any gesture, the gesture is considered completely ended.
We note that this amount of windows correspond to $0.25$ seconds given the acquisition frame rate of the \textit{Leap Motion} device (25 fps). This state improves the stability of the end gesture detection.
The method and the results have been computed using an Intel(R) Core(TM) i7-7700K CPU @ 4.20GHz CPU and an Nvidia GTX 1080 Ti GPU.

\subsection{Group 2: Image Based Methods}
\subsubsection{Introduction}
The two methods proposed by the Group 2 team are based on the transformation into images of the skeleton data captured by the Leap Motion sensor and their use for the training of a Convolutional Neural Network (CNN). The raw data points of the hand skeleton are rendered using a custom 3D visualizer: the images that constitute the dataset are captured by projecting the 3D skeleton on the xy plane. This view plane corresponds to the view from above which therefore represents the hands in a "natural" way, as a person normally sees them, and is kept constant throughout the image generation phase and at inference time. This approach has the substantial advantage of not requiring any labour-intensive feature engineering phase, so for the addition of new dynamic hand gestures to the vocabulary, the generation of the corresponding images and the retraining of the network is sufficient.

In order to provide temporal information to the network, the recent history of the gesture is represented by the fingertips traces of the hand. For the construction of the training set both the gestures of the SHREC '21 dataset and the gestures in the SFINGE3D dataset were used.

The first method used was described in detail in \cite{sfinge3d}: this method, which involves the use of a ResNet-50 for the classification of the rendered images, was retrained on the overall dataset exactly as described in the paper. In the inference phase, the same confidence thresholds used in the paper were also adopted. As shown in \cite{sfinge3d}, the disadvantage of this method is that the entire temporal dimension of the gesture is condensed into the two-dimensional image in the form of fingertip traces. Although this is sufficient for the inference of some types of gestures\footnote{The overall accuracy of the method on the \textit{SFINGE3D dataset} was 0.74.}, it is not for others, especially for those of typically very short duration or which develop mainly along the $z$ axis (\textit{grab}, \textit{pinch}, \textit{tap}) and therefore leave a minimal trace on the $xy$ plane. Another disadvantage of this method is that the recognition is not continuous, but typically occurs every 300 frames (parameter to be fine-tuned depending on the dataset) to limit the number of false positives.

The second method proposed was therefore fundamentally conceived to compensate for the shortcomings of the first method: in particular, the lack of a true temporal dimension in the first method of inference led us to the training of a ResNet-3D \cite{kenshohara2018spatiotemporal}, therefore able to perform 3D convolutions on volumes consisting of sequences of two-dimensional images.
As for the previous method, a significant offline data augmentation phase has been carried out for this method: during the generation of the training sequences, the images of the hands and fingertip traces were truncated in different moments in time to make the gestures incomplete. The continuous skeleton sequences were also sampled every 2 or 5 frames to produce the rendered images: this sampling was chosen to "compress" the longer gestures (some longer than 200 frames) into a shorter time span and as a further data augmentation. Noise was also added indipendently in the single rendered image, in the form of points around the skeleton of the hand to simulate residual traces of gestures.


This further data augmentation step has been added not only to help the ResNet-3D 50 converge to a robust solution\footnote{This kind of network has 46.4 million parameters \cite{leong2020semi}, almost double their 2D counterpart.} but also to try to better exploit the large model capacity of this class of networks. 

This offline data augmentation phase led to the generation of 72,720 image sequences (saved in WEBP format to optimize for space) that were randomly divided with an 80\%/20\% split to form training and validation sets. This phase is separate from the standard online data augmentation step that occurs during network training (resize, crop, rotate, warp, brightness, contrast, saturation).

The sequence length was set to 10 frames\footnote{With $seq\_len = 10$, the network was trained with tensors of type $torch.Size([bs, ch=3, seq\_len=10, h, w])$} and sequences shorter than $seq\_len$ frames were padded with black frames, while image sequences longer than $seq\_len$ were sampled randomly to perform further online data augmentation.

Network training was performed using the popular \textit{Fast.ai v2} library\cite{fastai} based on Pytorch, and the progressive resizing \cite{cellular_super_resolution} technique to optimize network convergence times. Using this technique, the training was carried out on images scaled progressively to $\frac{1}{8}$, $\frac{1}{6}$, $\frac{1}{4}$, $\frac{1}{3}$ of the original resolution of 1920x1080, for each training round on the network. All training rounds of the network took place in \textit{"frozen"} mode, thus training only the 19-neuron (untrained) output layer of the network for 1 epoch. For the last round, the network was also trained in \textit{"unfrozen"} mode, thus training all the layers of the network for a total of 7 epochs.
The optimizer used was Adam and as the loss function we chose LabelSmoothingCrossEntropy. LabelSmoothingCrossEntropy is defined as:
\begin{equation}
ls_{loss} = ( 1 −  \epsilon ) \xi(i)+\epsilon\sum \frac{\xi(j)}{N}
\end{equation}

where $\xi(j)$ is cross-entropy of x and i is the correct class. Through LabelSmoothingCrossEntropy we try to compensate for noisy labels in the training set: instead of wanting the model to predict 1 for the correct class and 0 for all the others, we teach it to predict $1 - \epsilon$ for the correct class and ε for all the others, being $\epsilon$ a small positive constant and N the number of classes of the problem.

The training took place on a GPU node of the new high-performance EOS cluster located within the University of Pavia. This node has a dual Intel Xeon Gold 6130 processor (16 cores, 32 threads each) with 128 GB RAM and 2 Nvidia V100 GPUs with 32 GB VRAM each. The training took place using PyTorch's DataParallel mode, so the batch size for the different training rounds was set to $32$, $16$, $8$, $4$ respectively to best occupy all the 64 GB of VRAM available on the two GPUs. The learning rate was set to $1\mathrm{e}{-2}$, $1\mathrm{e}{-2}$, $1\mathrm{e}{-2}$, $1\mathrm{e}{-4}$ respectively for the \textit{"frozen"} training rounds and $4\mathrm{e}{-4}$ for the 7 \textit{"unfrozen"} epochs in the last round of training. The final accuracy of the model used for the submission was 0.963 against the validation set.

Unfortunately, although these methods are promising in terms of approach to the problem, the short duration of the contest did not allow the team to optimize the results. As for the first method, it would probably have been sufficient to slightly lower the confidence threshold to consider a gesture as recognized to raise the positive detection score (possibly to the detriment of the false positive score which is still sufficiently low). The main problem of the second method, on the other hand, is certainly the too short sequence length: during the training the spatial resolution was privileged to the detriment of the temporal one and this did not allow the network to have sufficient context to learn. It is necessary to explore the trade-offs between spatial and temporal resolution to verify under which conditions the network learns best.

\subsection{Group 3: uDeepGRU and TSGR}
\subsubsection{Introduction}

For this track, Group 3 submitted results obtained from two different methods: the improved \textit{uDeepGRU} model~\cite{udeepgru,maghoumi2020dissertation} as well as a novel method dubbed \textit{Temporal Shift Gesture Recognizer} (TSGR). Both of their methods are based on deep neural networks. In the following they provide an overview of each one.

\subsubsection{uDeepGRU} 
\begin{figure}[t]
	\centering
	\includegraphics[width=\columnwidth]{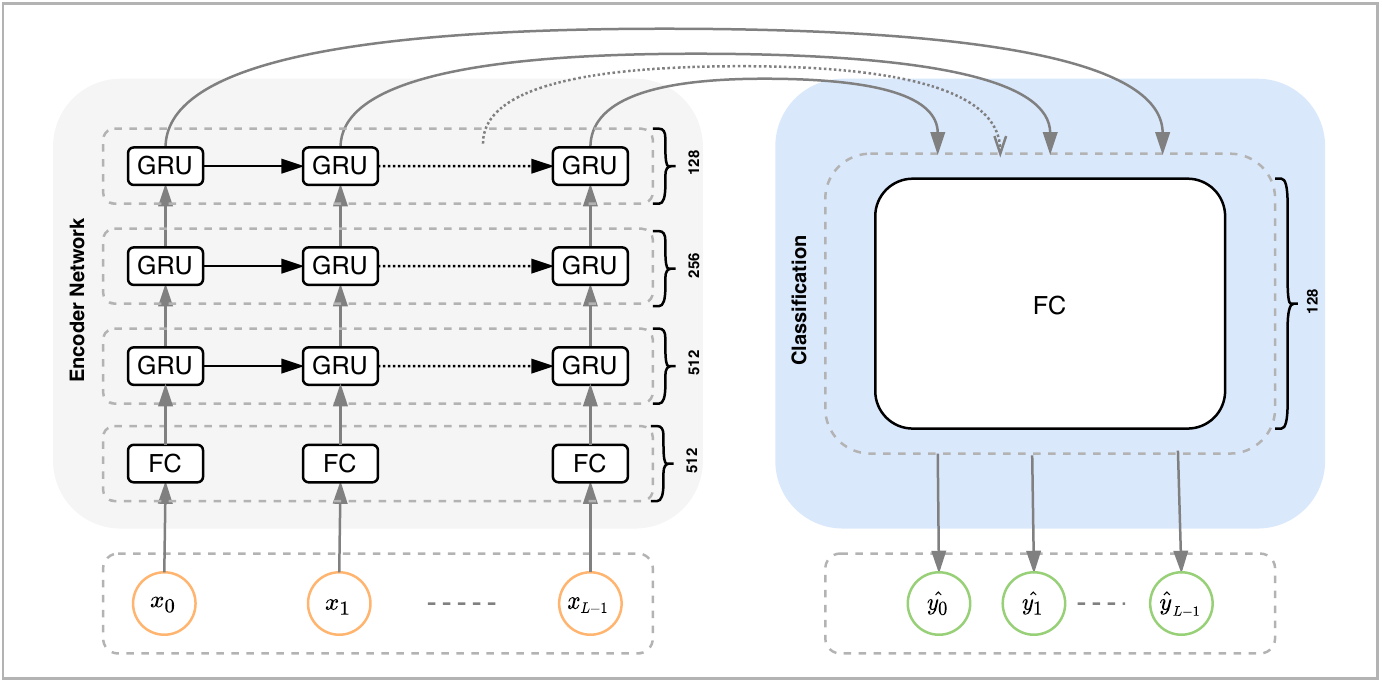}
	\caption{The improved uDeepGRU architecture which consists of an encoder network and a classification subnetwork.}
	\label{fig-udeepgru}
\end{figure}

Group 3 introduced the improved version of uDeepGRU \cite{maghoumi2020dissertation}. Figure~\ref{fig-udeepgru} depicts the network architecture of the improved model.

The uDeepGRU model is based on recurrent neural networks (RNN) and uses unidirectional gated recurrent units (GRU) as its main building block. Frames of a gesture sequence are sequentially fed to the network, and the network outputs the predicted label for every frame. Concretely, the network takes as input the feature vector $x_t$ at time step $t$ and produces the output label $\widehat{y_{t}} \in \{ \text{None} \}~\cup C$, where $C$ is the set of all possible gestures in the dictionary and \textit{None} indicates a no gesture. The the transition equation for each GRU cell in the uDeepGRU model is defined as:

\begin{align}
r_t &= ~~\sigma~~ \Big( \big( W_x^r~x_t + b_x^r \big) ~~+~~ \big(  W_h^r~h_{(t-1)} + b_h^r  \big) \Big) ~~~~\label{eq-gru}    \\
\begin{split}
\begin{split}
u_t &= ~~\sigma~~ \Big( \big( W_x^u~x_t + b_x^u \big) ~~+~~ \big(  W_h^u~h_{(t-1)} + b_h^u  \big) \Big) \nonumber    \\
c_t &= \text{tanh} \Big(  \big( W_x^c~x_t + b_x^c \big) ~~+~~  r_t\big( W_h^c~h_{(t-1)} + b_h^c \big) \Big)   \\
h_t &= u_t \circ h_{(t-1)} ~~+~~ \Big(1-u_t\Big) \circ c_t
\end{split}
\end{split}
\end{align}

\noindent
where $\sigma$ is the sigmoid function, $\circ$ denotes the Hadamard product, $r_t$, $u_t$ and $c_t$ are reset, update and candidate gates respectively and $W_p^q$ and $b_p^q$ are the trainable weights and biases. The initial hidden state tensor $h_0$ of all the GRUs in our model is initialized to zero. 

The improved uDeepGRU model introduces two main changes compared to the original implementation in~\cite{udeepgru}. First, there is an extra feature extraction layer at the beginning of the model. This layer consists of a fully-connected (FC) layer with $tanh()$ activation. Given $x_t$, the feature vector of an input frame at time step $t$, this layer computes $f_t = tanh\big(x_tW^T + b\big)$, where $W$ and $b$ are trainable weights and the bias term respectively. The goal of this feature extraction layer is to increase the feature extraction capacity of the model. Second, the training objective function for this model is the focal loss function~\cite{lin2017focal}. This loss function was originally proposed for problems with unbalanced labeled data. The online gesture recognition problem is an example of such unbalanced data, as many frames obtained from an input device are typically non-gestural interactions. Focal loss attempts to dynamically weight the standard cross-entropy loss to dampen the effect of well-classified examples on the final loss value. For a model which outputs $p$ as the probability of a given class label, the focal loss is defined as:

\begin{equation}
\label{eq-focal-loss}
\text{L}_{\text{~focal}} = -(1-p)^{\gamma}~\text{log}(p)
\end{equation}

\noindent
where $\gamma \ge 0$ determines how much emphasis is put on misclassified examples. We use $\gamma=1$ in our implementation and train uDeepGRU end-to-end.

\subsubsection{TSGR}

Although RNN-based models have demonstrated great capabilities in sequence modeling and prediction tasks, they suffer from a few critical drawbacks. Namely, RNNs rely on their hidden state tensors for modeling the relationships across different time steps of their inputs. Also, most RNN-based models suffer from problems such as vanishing or exploding gradients which make them difficult to train.

To address these issues, Lin~\textit{et al.}~\cite{lin2019tsm} recently introduced the \textit{Temporal Shift Module} (TSM) an alternative sequence modeling paradigm using the temporal shifting of input features. The basic idea behind such methods is simple, yet powerful: at each time step during the processing of a temporal sequence, replace a portion of the features of the current time step with those of other time steps via shifting those features across the temporal dimension. This allows the network to perform temporal modeling across extracted features. Depending on the task, the shifting operation can either be bidirectional or unidirectional. Unidirectional shifting is suitable for online recognition tasks as during recognition, only the features of the prior time steps are available.

\begin{figure}[t]
	\centering
	\includegraphics[width=\columnwidth]{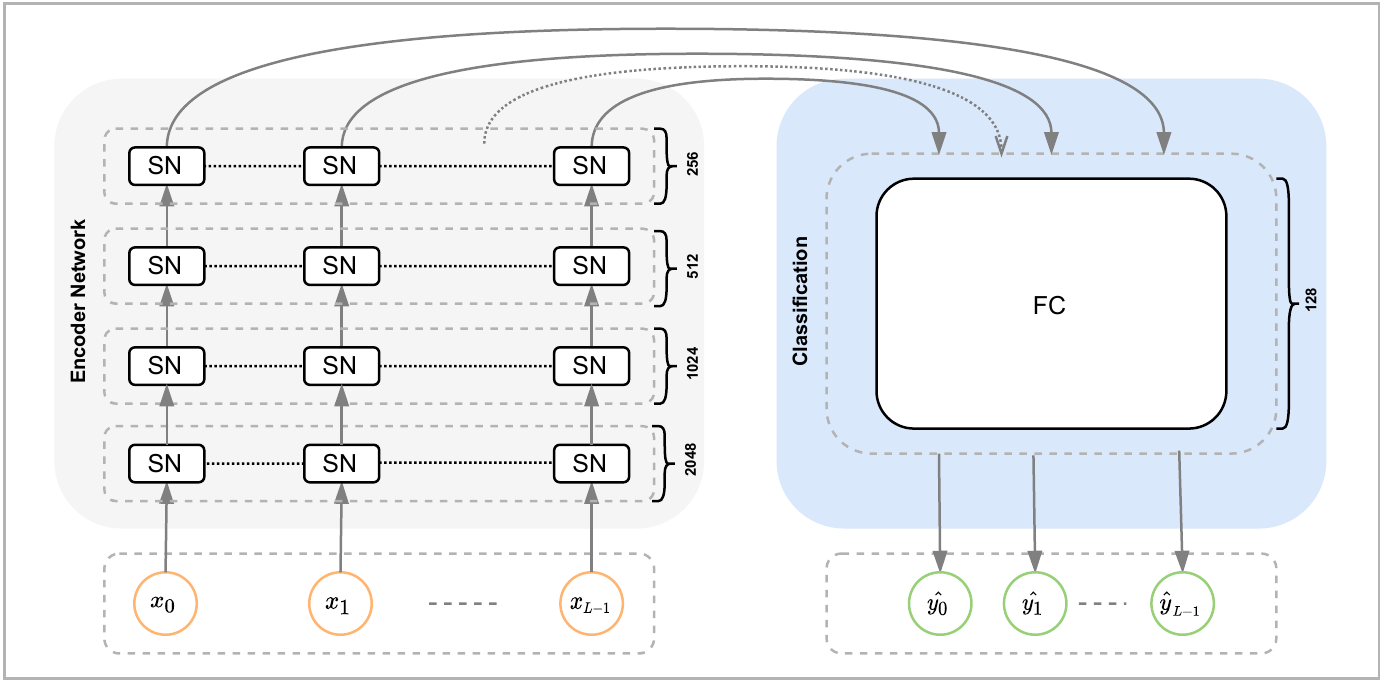}
	\caption{The TSGR architecture which consists of an encoder network and a classification subnetwork. Each shift node (SN) has access only to the features of the previous time steps in that same layer.}
	\label{fig-tsgr}
\end{figure}

Based on this idea Group 3 devised TSGR, their second online gesture recognizer. The TSGR model is depicted in Figure~\ref{fig-tsgr}. Similar to uDeepGRU, our TSGR model takes the input features for every time step and produces the output class prediction. This model, which is conceptually simpler than uDeepGRU, consists of only FC layers with TSM layers in between. Each TSM layer replaces half of the features of the current frame with those of the past 5$^{\text{th}}$ frame. This implies that the amount of feature shifting in our model is five frames. Although the original TSM model~\cite{lin2019tsm} shifts the features by one, Group 3 experimentally found five shifts to work better for this track's data.

We refer to the combination of FC and TSM layers as \textit{shift nodes} (SN) henceforth. Each SN consists of a TSM layer along with two FC layers, namely FC$_{\text{Shift}}$ and FC$_{\text{Residual}}$ with $ReLU()$ activations\footnote{Except for the very first layer, where $tanh()$ is used}. The dimensionality of both FC layers is the same, however FC$_{\text{Residual}}$ layers do not include a bias term.
The sequence of operations inside each node is as follows. At each time step $t$, the node takes the feature vector $f_t$, performs the shifting operation by replacing half of the values in $f_t$ by those computed for $f_{t-5}$. The node then passes these features through FC$_{\text{Shift}}$ and saves the intermediate results $f^{\text{Shift}}_t$. The node also computes and stores the intermediate result $f^{\text{Residual}}_t = \text{FC}_{\text{Residual}} \big( f_t \big)$. The final output of the node is computed as $f^{\text{SN}}_t = f^{\text{Shift}}_t + f^{\text{Residual}}_t$. Additionally batch normalization~\cite{batch-norm} is applied before the final result is passed to the next layer. Similar to uDeepGRU, we optimize the focal loss function to train TSGR end-to-end. 

\subsubsection{Implementation and Training}
Each frame of the data is treated as one 60-dimensional vector $x_t$ obtained by concatenating the 3D position of all joints. Every feature vector was z-score normalized using the mean and standard deviation of all feature vectors in the training set.

Group 3 implemented their models in PyTorch. Both models are trained end-to-end on the training set, with six random sequences withheld for validation. Training was done using the Adam optimizer~\cite{kingma2014adam} with a learning rate of 0.0002 and a mini-batch size of 10. The maximum length of a training sample was fixed to 256 (longer samples were split to chunks of at most 256 frames). Counter-intuitively, they found any kind of data augmentation on the training set to be harmful. After training concluded, we saved the model that produced the best F$_1$ score on the validation set. 

\subsubsection{Results}
\label{sec:results}
At test time, Group 3 runs each test sample through the network and obtain per-frame class labels $\hat{y}_t$. They do not perform any post-processing on the output results and the hardware used for the compute the results is a AMD Ryzen 3900x processor and
an NVIDIA Titan RTX GPU with 24 GB VRAM.

They obtained three sets of results. The first set consists of the results of an ensemble of 10 uDeepGRU models, each trained on a different portion of the training set. The second set is the classification labels obtained from a single TSGR trained model, and the last set contains the results of an ensemble of 15 TSGR models, each trained on a different portion of the training set.

\subsection{Group 4: Spatial-Temporal Graph Convolutional Network}
\subsubsection{Data preprocessing}
Based on the task description, the dataset is mixed with random hand movements between real gestures. Group 4 added those noisy segments labeled as non-gestures into the training procedure to improve robustness of the model. Next, they extract both gestures and non-gestures data with a length of 200 frames for each segment. Due to the limitation on training data, a stratified 5-fold based on class-distribution is applied to avoid under and over fitting.  

\subsubsection{Classification Model: Spatial-Temporal Graph Convolutional Networks}
\begin{figure}[h]
	\centering
	\includegraphics[width=1.2\columnwidth]{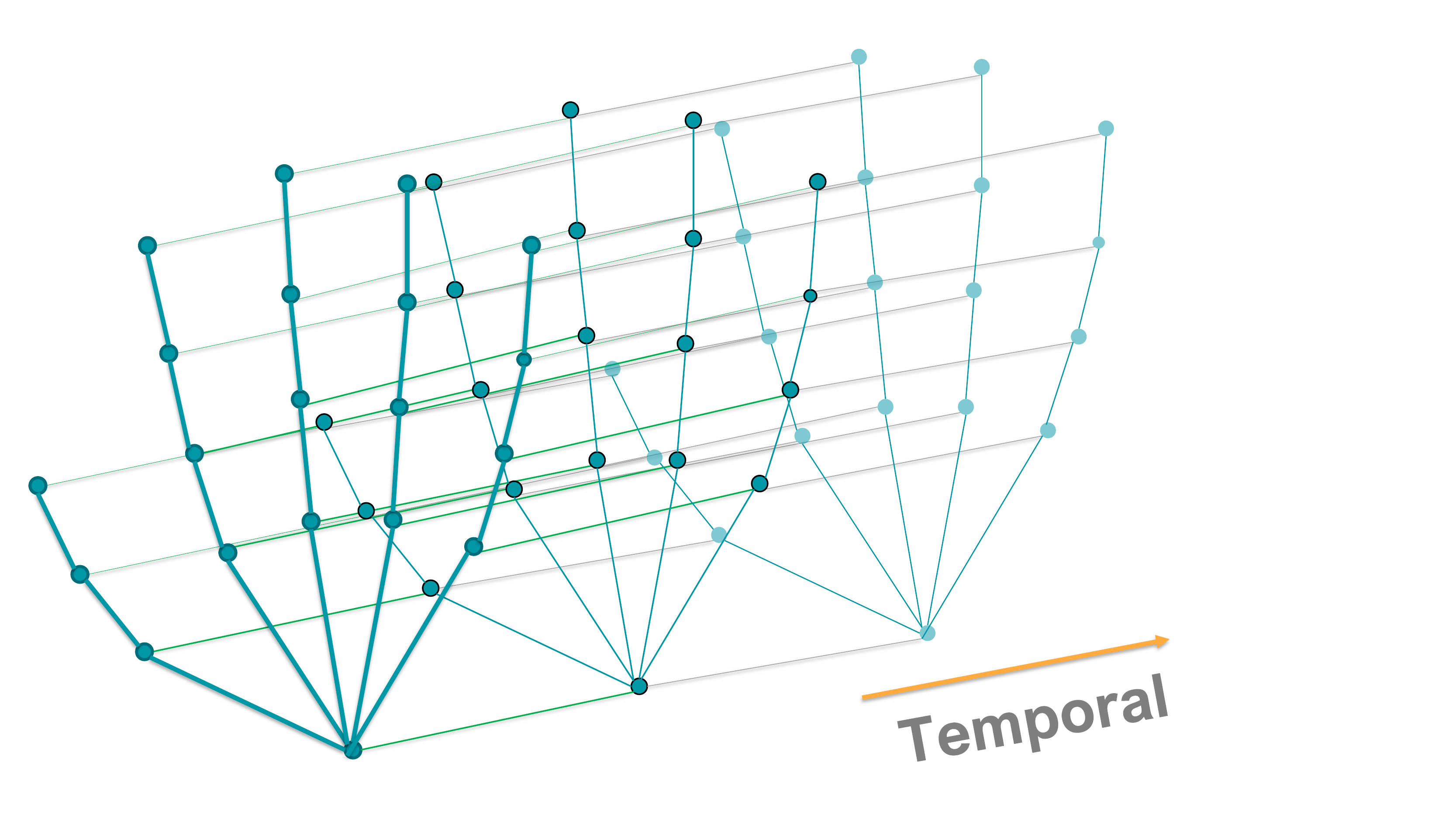}
	\caption{Spatial-temporal graph neural networks for hand gesture recognition}
	\label{fig:st_gcn}
\end{figure}

 Spatial-Temporal Graph Convolutional Networks
 (ST-GCN) \cite{st_gcn} is an extended version of graph neural networks to a spatial-temporal graph model. 
 The graph can learn patterns embedded in the spatial configuration by exploring locality of graph convolution as well as temporal dynamics. As proposed by Yan et al., Group 4 constructs a sequence of skeleton graphs, each node represents a joint of the hand. Moreover, there are spatial edges for building up the connectivity of joints according to natural structure of human hands  and temporal edges connecting the same joints across continuous frames of actions, described as Figure \ref{fig:st_gcn}. Also, each node has its own features composed of 3D coordinates and quaternions. In the classification module using ST-GCN, features of a joint at frame t is a vector with length 7 consist of $[x, y, z, a, b, c, d]$
where $(x,y,z)$ expresses 3D coordinates and a quaternion q is described by $q = a + bi + cj + dk$.\\
In the detection module, we only use $(x,y,z)$ to indicate potential candidates. 

\subsubsection{Energy-based Detection Module}
With respect to detection and localization module, Group 4 uses a non-deep learning approach by leveraging the shift of every joint of human hand over sequences of consecutive frames. They define an energy-based function to calculate the amount of energy accumulated in a 
window of length L inside the gesture stream: 
\begin{equation}
\scalebox{0.9}{
    $\begin{aligned}
    E(w) = \sum_{n=1}^{N}\sum_{t=1}^{L}\sqrt{(\frac{w_{n,t}^{x}}{w_{n,t-1}^{x}} - 1)^2 + (\frac{w_{n,t}^{y}}{w_{n,t-1}^{y}} - 1)^2 + (\frac{w_{n,t}^{z}}{w_{n,t-1}^{z}} - 1)^2} 
    \end{aligned}
    $
    }
    \label{energy}
\end{equation}

Index $t$ ranges from 1 to L. $n$ is the index of the hand joint ranging from 1 to N. So, $w_{n,t}^x$ represents the x-coordinate of $n^{th}$ joint at frame $t$. \\

Using Equation \ref{energy}, Group 4 adopted a sliding window approach, estimating the value of $E(w_i)$ on multiple windows $w_i$ of length L starting at different locations, with consecutive windows $w_i$, $w_{i+1}$ separated by a fixed 
stride step.

Authors then take as candidate gestures all the windows corresponding to local maxima of the energy, e.g. those satisfying the following conditions:
\begin{eqnarray}
    \delta E(w_{i}) & <  & \epsilon, \\
    \delta E(w_{i-1}) & >  & 0, \nonumber \\
    \delta E(w_{i+1}) & <   & 0 \nonumber
    \end{eqnarray}
$\delta E(w_i)$ is the time derivative of energy of segment $w_i$, approximated by:
\begin{equation}
    \delta E(w_i) = \frac{E(w_{i+1}) - E(w_{i-1})}{(i+1)-(i-1)}
\end{equation}

After detecting possible segments, Group 4 feeds those candidates into the ST-GCN model. Those segments that are predicted to belong to a non-gesture class with confidence score $>$ threshold $\alpha$ or be a member of gesture classes with confidence score $<$ threshold $\beta$ are filtered. Threshold $\alpha$ and $\beta$ are chosen based on the validation data.

\subsubsection{Trajectory-based fine-tuning with PCA and gradient histogram}
\begin{figure}[h]
	\centering
	\includegraphics[width=1.0\columnwidth]{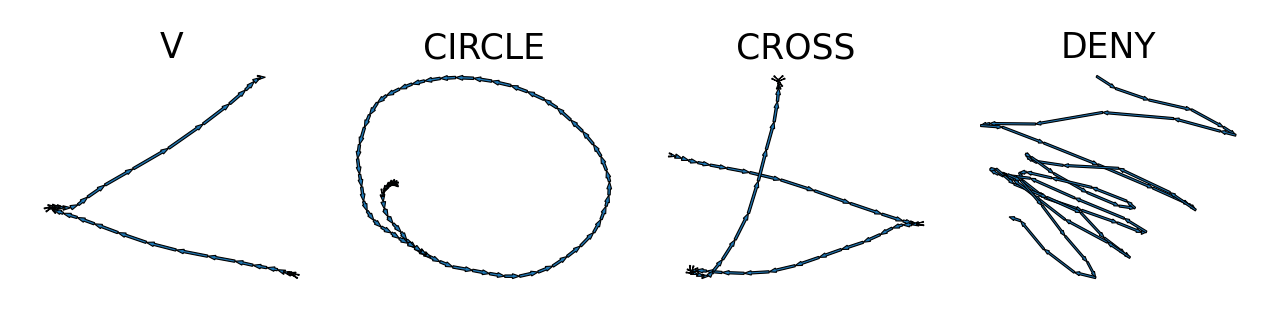}
	\caption{2D Trajectories of hand gestures}
	\label{fig:traj}
\end{figure}

Group 4 noticed that some gestures can be easily classified by leveraging their trajectory. Among the 20 joints of a human hand in the dataset, they chose IndexEnd of an index finger to consider action's trajectory of such gestures: CIRCLE, V, CROSS, DENY. The principal component analysis is applied to reduce 3D coordinate to 2D system, see figure \ref{fig:traj}. 

\begin{figure}[h]
	\centering
	\includegraphics[width=1.05\columnwidth]{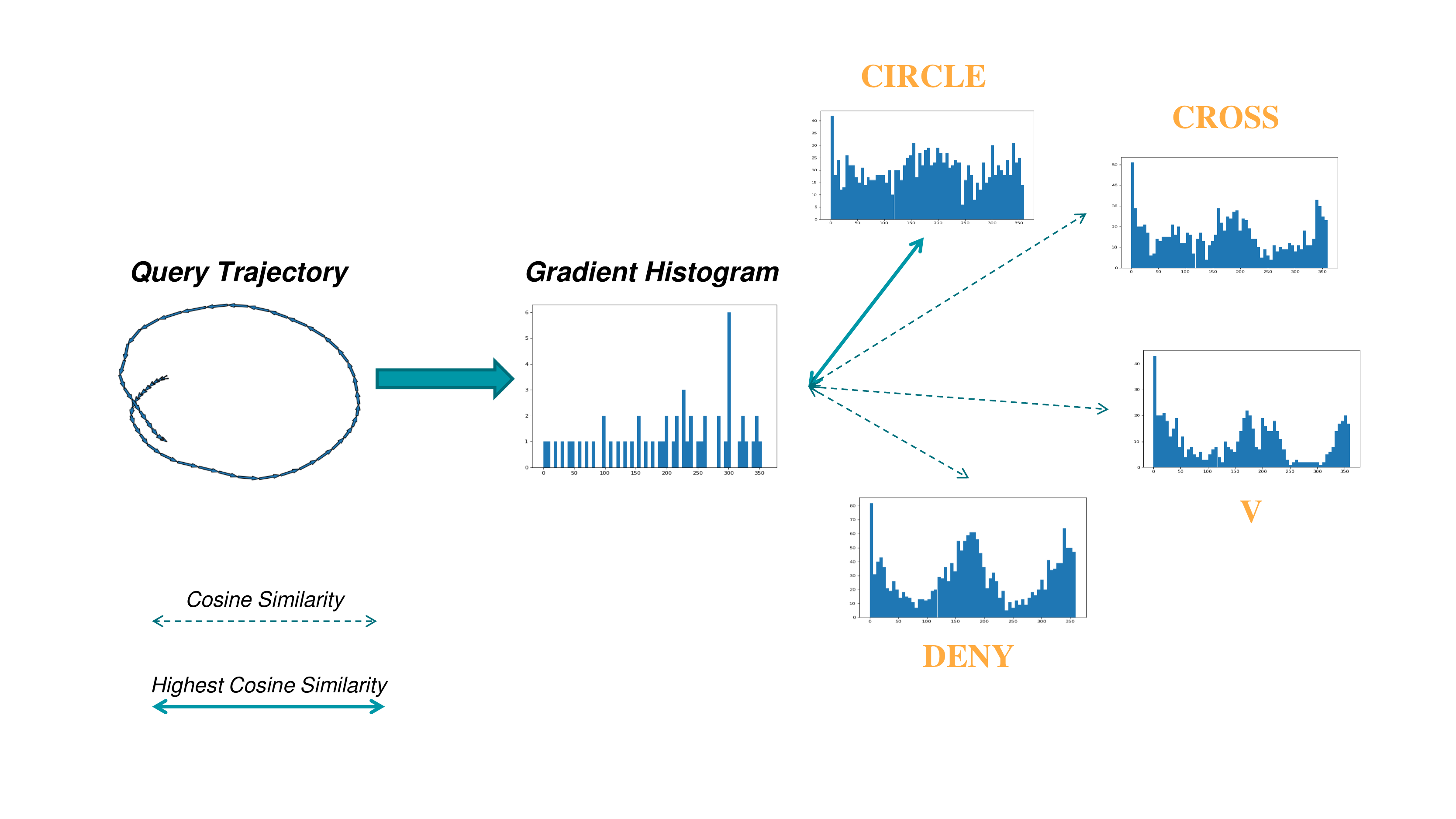}
	\caption{Cosine similarity between gradient histograms}
	\label{fig:pca_traj}
\end{figure}

Next, those sequences of 2D coordinates are utilized to find gradient vectors with Ox and Oy axis and their angles. A histogram with N bins ranging from $-\pi/2$ to $\pi/2$ is computed on the frequency of calculated angles \cite{thompson_shrec20}. This 1D orientation histogram is viewed as a feature vector of each gesture. With regard to the training set, every class of gestures comprises some segment members that belong to this class. Thus, each class of gesture is represented by the mean histogram feature vector of its members. In the inference phase, they extract the gradient histogram feature vector of every candidate segment and then compare it with each class’s representation vector using Cosine Similarity, described as Figure \ref{fig:pca_traj}. Finally, the highest similarity score is ensembled with confidence score from the ST-GCN model and the predicted label is returned.

\subsubsection{Experiments}
Group 4 did experiments on single ST-GCN models, k-fold models, and ensemble models as well. Finally, they chose 3 configurations of them that correspond to 3 RUNS.

\begin{itemize}
    \item RUN1: Single ST-GCN model.\\
    (CPU Intel core i5-8300H 2.3GHz with Turbo Boost up to 4.0GHz)
    \item RUN2: ST-GCN model with stratified 5-fold.\\
    (GPU Tesla P100 16GB)
    \item RUN3: 5-fold ST-GCN models ensemble with gradient histogram.\\
    (CPU Intel Core i5-8300H 2.3GHz with Turbo Boost up to 4.0GHz)
\end{itemize}


\section{Evaluation Results}
A summary of the results for each group, averaged over all the gestures, is presented in Table \ref{tab:summary}. The table also shows the results of the execution time of the methods by reporting the Total Time and the Classification Time. The first is a measure of the time each method takes to compute results for the entire test set (i.e. all the sequences), the second, is a measure of the average time needed, for each method, to perform a single gesture classification.
Times show that all the methods are suitable for real-time applications, even if the performances are not directly comparable as the software has been executed on different architectures.

\begin{figure*}[ht]
	\centering
	\includegraphics[width=1\linewidth]{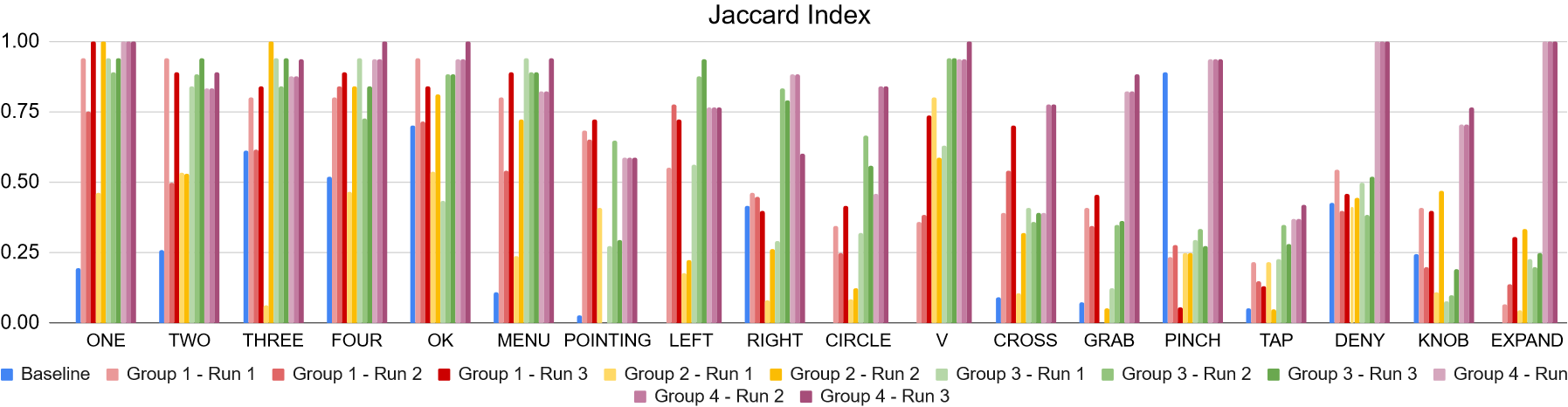}
	\caption{Jaccard index per class on all the test sequences.}
	\label{fig:jaccard}

	\centering
	\includegraphics[width=1\linewidth]{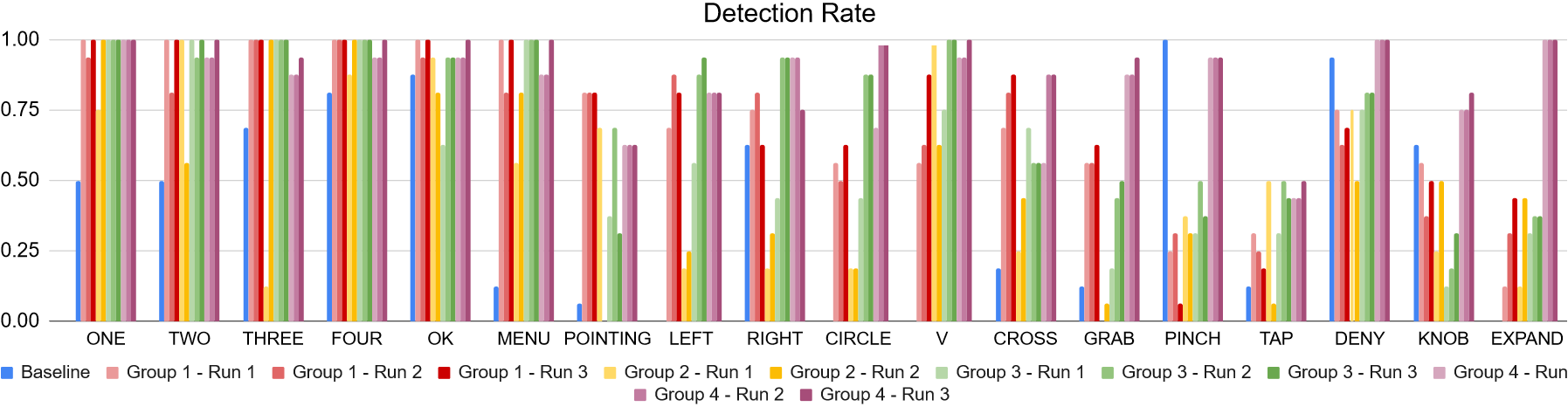}
	\caption{Detection rate per class on all the test sequences.}
	\label{fig:detection}

	\centering
	\includegraphics[width=1\linewidth]{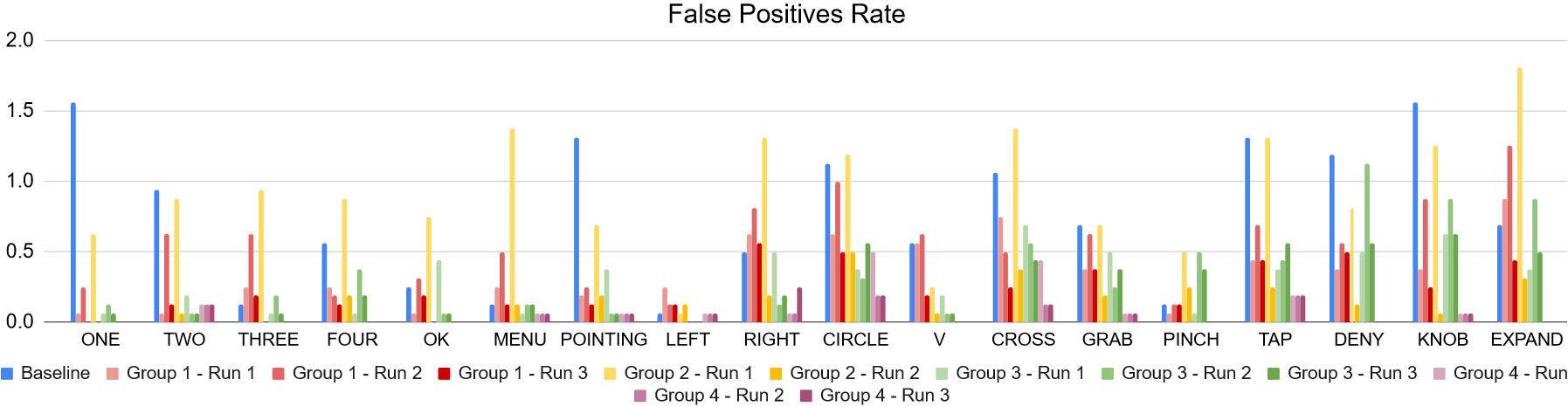}
	\caption{False Positives rate per class on all the test sequences.}
	\label{fig:fp}
\end{figure*}

\begin{table}[]
\centering
\resizebox{\columnwidth}{!}{%
\begin{tabular}{|c|c|c|c|c|c|}
\hline
\rowcolor[HTML]{C0C0C0} 
\textbf{Method} & \textbf{Det. Rate} & \textbf{FP Rate} & \textbf{Jac. Ind.} & \textbf{Tot.Time(s)} & \textbf{Class.Time(s)} \\ \hline
\rowcolor[HTML]{DAE8FC} 
Baseline        & 0.3993 & 0.7639 & 0.2566 &    1161.0    &    0.08      \\ \hline
\rowcolor[HTML]{FFFFFF} 
Group 1 - Run 1 & 0.7014 & 0.3576 & 0.5501 & 435.5 & 1.36    \\ \hline
\rowcolor[HTML]{FFFFFF} 
Group 1 - Run 2 & 0.6875 & 0.5521 & 0.4737 & 435.5 & 1.36    \\ \hline
Group 1 - Run 3 & 0.7292 & 0.2569 & 0.6029 & 435.5 & 1.36    \\ \hline
\rowcolor[HTML]{DAE8FC} 
Group 2 - Run 1 & 0.4861 & 0.9271 & 0.2772 & 48781.0 & 0.41   \\ \hline
\rowcolor[HTML]{DAE8FC} 
Group 2 - Run 2 & 0.4931 & 0.1667 & 0.4458 & 4897.2   & 0.81    \\ \hline
\rowcolor[HTML]{FFFFFF} 
Group 3 - Run 1 & 0.6042 & 0.3021 & 0.4987 & 66.7 & \num{0.6e-4}   \\ \hline
Group 3 - Run 2 & 0.7569 & 0.3403 & 0.6194 & \textbf{0.4}  & \textbf{\num{0.3e-5}} \\ \hline
\rowcolor[HTML]{FFFFFF} 
Group 3 - Run 3 & 0.7431 & 0.2708 & 0.6238 & 5.1  & \num{0.4e-4} \\ \hline
\rowcolor[HTML]{DAE8FC} 
Group 4 - Run 1 & 0.8403 & 0.0903 & 0.7925 & 94.6    & \num{0.6e-2}        \\ \hline
\rowcolor[HTML]{DAE8FC} 
Group 4 - Run 2 & 0.8750 & \textbf{0.0556} & 0.8353 & 281.4   & 0.03    \\ \hline
\rowcolor[HTML]{DAE8FC} 
Group 4 - Run 3 & \textbf{0.8993} & 0.066  & \textbf{0.8526} & 289.2  & 0.16         \\ \hline
\end{tabular}%
}
\caption{Average scores obtained by all the proposed techniques on the test sequences and corresponding execution times for the classification of all the test data and for a single prediction at a given sequence point (online detection). Bold fonts indicate best results.}
\label{tab:summary}
\end{table}

The methods based on ST-GCN provide clearly the best performance, and it is interesting also to note that the combination of the network based method with simple heuristics is able to improve the scores. These methods also provide a low number of false positives, that are instead not negligible in the other methods.

To better understand the outcomes of the different techniques it is useful to analyze the scores related to the single gesture classes.

The bar charts in Figures \ref{fig:jaccard}, \ref{fig:detection} and \ref{fig:fp} show the per-class scores of all the methods.

As expected, all the methods present good performances for the static gestures (i.e. from ONE to POINTING). The average JI for these gestures is  $0.73$. However, some methods result in a non negligible number of false positives that could make difficult to use the techniques in a practical scenario. The average false detection rate on static gestures is, in fact, $24\%$, meaning a false positive detected every 4 gesture recognized, that is quite high for practical purposes.

The POINTING gesture is the most challenging static one, and it is reasonable, being the one with more relevant variations in the execution.

\begin{figure}[h]
	\centering
	\includegraphics[width=1.0\columnwidth]{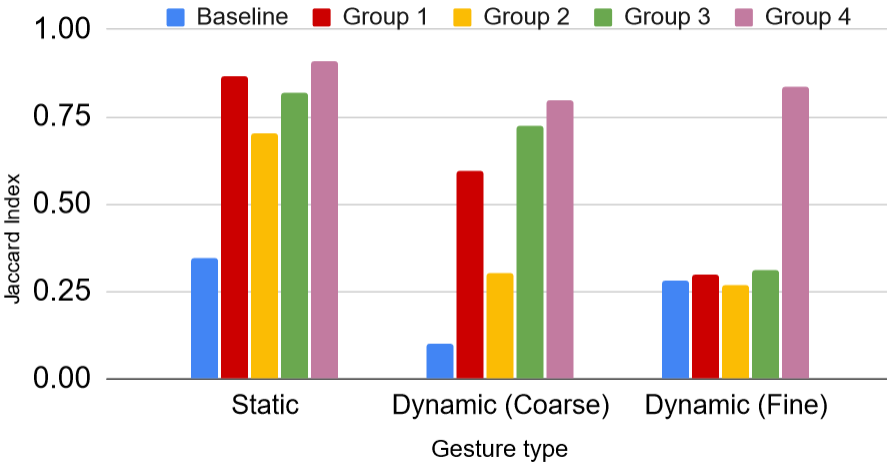}
	\caption{Jaccard Index by gesture type (best run for each group)}
	\label{fig:gesturetype}
\end{figure}

Dynamic gestures are much harder to recognize 
(average JI $0.46$, average FD $0.41$), with the notable exception of the ST-GCN based methods. As shown in Figure \ref{fig:gesturetype}, considering the best run (i.e. highest Jaccard Index) for each group, the results show that among the dynamic gestures, those considered fine and therefore characterized also by the single fingers' trajectories (GRAB, PINCH, TAP, DENY, KNOB, EXPAND), present further issues for most of the methods.
ST-GCN works well on most of the gesture class with just a few exceptions (TAP, POINTING, KNOB).

Some surprising facts appear from the analysis of single classes: for example the LEFT gesture presents a very low number of false positives despite being a dynamic one, while 
the RIGHT gesture is found in a lot of false detections.

The PINCH gesture is surprisingly detected easily and with few false detections by the baseline method, and it is hardly detected by most of the network based techniques.

\section{Discussion}
The evaluation outcomes provide useful insights for the design of online gesture recognizers usable "in the wild".
The techniques tested provide promising scores given a limited number of annotated sequences for training.
Given the short amount of time available for the contest this is a good result.

A nice aspect of the submissions received is that the proposed methods are quite different from each other and are exemplars of the principal network-based approaches proposed in the literature for these tasks, namely Recurrent Networks, Graph Networks, Transformer Networks, and Convolutional Neural Networks.
The fact that Spatiotemporal Graph Convolutional networks provide the best results is consistent with the literature on action recognition from body skeletons, where the best scores on the related benchmarks have been obtained with similar approaches. 
However, on hand gesture recognition, good results on old benchmarks have been obtained with modified versions of LSTM including neighbor information \cite{min2020efficient} or using 1D CNN \cite{yang2019make}. These methods have not been proposed in this contest, and we plan to test them on our data as future work.

It must be noted, then, that the performances of the different techniques depend on hyperparameters tuning and training data augmentation, and, given the short amount of time available for the contest, the results could be improved and the ranking changed.

In any case, it is worth noting that for all the proposed methods but the ST-GCN-based dynamic gestures are not well handled and false positives are a relevant issue.

With all the methods, selected gestures were hard to be detected (e.g. TAP, KNOB, POINTING). 
A possible future research direction is therefore to investigate the reasons for these problems, which may rely on similarities between segments of different gestures or the variability in the execution. 
The goal could be to create optimal dictionaries for gestural interfaces avoiding the inclusion of "problematic" classes.

The improvements of the ST-GCN results obtained with the addition of simple handcrafted similarity evaluations and the fact that simple handcrafted features work well on specific gestures (PINCH) show that it is hard to have a generic method well-suited for the recognition of all the gesture types.
A viable option to address this issue could be to define multiple recognizers for specific gestures.

Problems with dynamic gestures could also derive from the limited numbers of training sequences and subjects performing the gestures and from the fact that the users performing the gestures in the test set were not involved in the recording of the training set. 
The availability of larger and more varied training sets could be exploited to increase the detection performances.
However, for the practical use of recognizers in interface design, the availability of gesture recognizers that can be trained with few examples would be particularly useful.

Another aspect that should be investigated is related to the computational load required for the online classification. While all the networks can be used for online recognition on a high-end PC, the possibility to have them running on Hololens or Oculus Quest needs to be checked.

We plan to update the dataset by adding new data. Furthermore, while the current gestures were recorded within a continuous gesticulation, but separated by non-gesture actions, we want to record sequences with series of adjacent gestures and design a new task involving the detection of series of atomic gestures.

\section{Conclusions}
The development of effective and flexible gesture recognizers able to detect and correctly classify hand gestures of different kinds is fundamental not only to enable advanced user interfaces for Virtual and Mixed Reality applications, but also, for example, to enable the realization of touchless interfaces like public kiosks, that are expected to replace touch-based ones after the emergence of the pandemic issues, being a more hygienic and safer solution.
It is, therefore, important to support the research on this kind of tool, developing benchmarks able to test the algorithms on realistic user scenarios.
The SHREC 2021: Track on Skeleton-based Hand Gesture Recognition in the Wild tries to do this. We believe that the dataset created and the methods proposed by the participants will be a useful asset for the researchers working on this topic.



\bibliographystyle{cas-model2-names}

\bibliography{cas-refs}


\end{document}